\documentclass[]{bytedance_seed}
\usepackage{tabularx}

\usepackage[toc,page,header]{appendix}

\usepackage{minitoc}
\usepackage{cleveref} 
\usepackage{subcaption}
\usepackage{booktabs}
\usepackage{graphicx}
\usepackage{pgfplots}
\usepackage{pgfplotstable}
\usepackage{xcolor}
\usepackage{CJKutf8}
\usetikzlibrary{patterns}
\usepackage{float}
\usepackage{caption}
\newcommand{\method}{Seed1.5-Thinking\xspace}

\usepackage{natbib}
\usepackage{latexsym}

\usepackage{url}
\usepackage{amssymb}
\usepackage[utf8]{inputenc}
\usepackage{microtype}
\usepackage{booktabs}
\usepackage{pifont} 
\usepackage{multirow}
\usepackage{makecell}
\usepackage{paralist}
\usepackage{xspace}
\usepackage{color}
\usepackage{xcolor}
\usepackage{colortbl}
\usepackage{adjustbox}
\usepackage{hyperref} 
\usepackage[edges]{forest}
\usepackage{tikz} 
\usepackage{caption}
\usepackage{amsfonts}

\hypersetup{
    colorlinks,
    linkcolor={blue!80!black},
    citecolor={blue!80!black},
}
\tikzset{
    root/.style =             {align=center, text width=1cm, rounded corners=3pt, line width=0.3mm, fill=gray!10, draw=gray!80, font=\small},
    demographic/.style =         {align=center, text width=1.8cm, rounded corners=3pt, line width=0.3mm, fill=blue!10, draw=blue!80, font=\footnotesize},
    demographic_work/.style =    {align=center, text width=10cm, rounded corners=3pt, line width=0.3mm, fill=blue!10, draw=blue!0, font=\footnotesize},
    character/.style =         {align=center, text width=1.8cm, rounded corners=3pt, line width=0.3mm, fill=red!10, draw=red!80, font=\footnotesize},
    character_work/.style =    {align=center, text width=10cm, rounded corners=3pt, line width=0.3mm, fill=red!10, draw=red!0, font=\footnotesize},
    personalization/.style =           {align=center, text width=1.8cm, rounded corners=3pt, line width=0.3mm, fill=cyan!10, draw=cyan!80, font=\footnotesize},
    personalization_work/.style =      {align=center, text width=10cm, rounded corners=3pt, line width=0.3mm, fill=cyan!10, draw=cyan!0, font=\footnotesize},
    risk/.style =         {align=center, text width=1.8cm, rounded corners=3pt, line width=0.3mm, fill=orange!10, draw=orange!80, font=\footnotesize},
    risk_work/.style =    {align=center, text width=10cm, rounded corners=3pt, line width=0.3mm, fill=orange!10, draw=orange!0, font=\footnotesize},
}

\usepackage{CJK}

\title{\method: Advancing Superb Reasoning Models with Reinforcement Learning }

\affiliation{ByteDance Seed}

\contribution{Full author list in Contributions}

\abstract{

We introduce \method, capable of reasoning through thinking before responding, resulting in improved performance on a wide range of benchmarks.
\method achieves $86.7$ on AIME $2024$, $55.0$ on Codeforces and $77.3$ on GPQA, demonstrating excellent reasoning abilities in STEM and coding.
Beyond reasoning tasks, the method demonstrates notable generalization across diverse domains. For instance, it surpasses DeepSeek R1 by 8\% in win rate on non-reasoning tasks, indicating its broader applicability.
Compared to other state-of-the-art reasoning models, \method is a Mixture-of-Experts (MoE) model with a relatively small size, featuring 20B activated and 200B total parameters.
As part of our effort to assess generalized reasoning, we develop two internal benchmarks, BeyondAIME and Codeforces, both of which will be publicly released to support future research. Model trial link: \url{https://www.volcengine.com/experience/ark}.

\date{April 10, 2025}
}

\begin{document}

\maketitle



    


\vspace{-20pt}
\begin{figure}[H]
    \centering
    \small
    \includegraphics[width=0.95\linewidth]{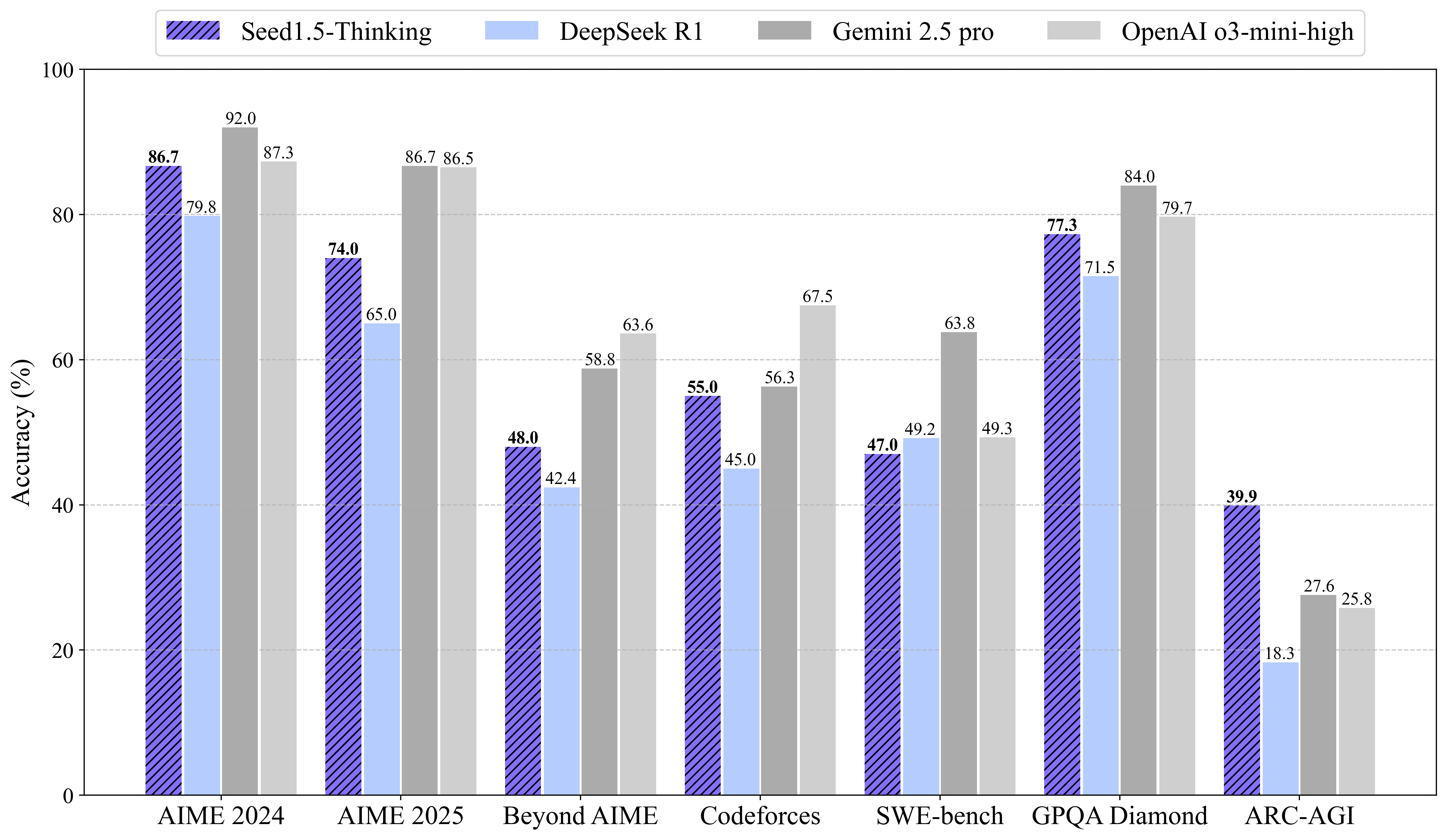}
    \vspace{-8pt}
    \caption{Benchmark performance on reasoning tasks}
    \label{fig:enter-label}
\end{figure}
\vspace{-10pt}

\section{Introduction}
 Driven by large-scale reinforcement learning on large language models, reasoning models have seen significant advancements. Notably, OpenAI's o1 series \citep{o1}, DeepSeek's R1 \citep{r1}, Google's Gemini 2.5 \citep{gemini2.5}, and Anthropic's Claude 3.7 \citep{claude3.7} have emerged as state-of-the-art models, each making substantial progress in logical reasoning, mathematical problem-solving, and code generation.
These advancements underscore a shift toward more structured, efficient and scalable reasoning models, with ongoing research focusing on  training efficiency, long chain-of-thought, and large-scale reinforcement learning. 

In this work, we present a new reasoning model, called \method. 
This model has achieved strong performance in both reasoning and non-reasoning tasks.
\begin{description}
\item[Mathematical Reasoning]: For math competition, \method achieves $86.7$ on AIME 2024, matching the performance of o3-mini-high and significantly outperforming o1 and DeepSeek R1, demonstrating competitive strength. 	Since AIME 2024 no longer provides sufficient discrimination, we construct a more challenging evaluation set named BeyondAIME. All problems in BeyondAIME are newly curated by human experts and designed to minimize the chance of being solved through memorization or guessing. While \method surpasses both o1 and R1, there remains a performance gap compared to o3 and Gemini pro 2.5. 	This also further demonstrates the discriminative power of the new evaluation set.
\item[Competitive Programming]: For the evaluation of competitive programming, we adopt Codeforces as our benchmark. 	Unlike some prior works that rely on Elo Scores, which contains estimation and are not directly comparable, we adopt a concrete evaluation protocol based on the most recent 12 Codeforces contests. Specifically, we report pass@1 and pass@8 metrics, where pass@k indicates whether the model solves the problem within k attempts, i.e., selecting the best result from k generated submissions. We choose to report pass@8 since it provides more stable results and aligns more closely with actual user submission patterns. \method outperforms DeepSeek R1 on both metrics, though a performance gap remains compared to o3. The evaluation set will be made publicly available in a future release.
\item [Science]: \method reaches a score of 77.3 on GPQA, close to o3-level performance. Importantly, this gain is largely attributed to improved generalization from mathematical training, rather than an increase in domain-specific science data.
\item [Non-reasoning Tasks]: For non-reasoning tasks, \method is evaluated using a test set designed to replicate real-world user needs. Through human evaluations conducted against DeepSeek R1 across diverse scenarios, \method demonstrates significant advancements: it attains an 8.0\% overall rise in users' positive feedback, thereby highlighting its augmented ability to manage intricate user scenarios.
\end{description}

There are three key points in the development of high-quality reasoning models: training data, RL algorithm, and RL infrastructure. We have devoted considerable effort to these three areas, and we will discuss them in detail.
\begin{description}
\item[Data]  For SFT training, unlike conventional post-training data, reasoning models rely on chain-of-thought data, which explicitly outlines the step-by-step reasoning process. Our preliminary experiments showed that too much non-CoT SFT data can significantly reduce the model's ability to explore. For RL training, we incorporate four categories of data: STEM problems, code-related tasks, logic reasoning and non-reasoning data like creative writing and dialogue. Among these, the logic reasoning data contributes to performance improvements on the ARC-AGI benchmark significantly.
The math data exhibits strong generalization capabilities and can lead to broad performance improvements across tasks.
\item[RL Algorithm] RL training of reasoning models is highly unstable and often crashes, especially for models without SFT. Sometimes, the score difference between two runs can be as high as 10 points. The stable training of RL systems is crucial for the success of reasoning models. 
To address these long-standing issues, we have pioneered VAPO\citep{vapo} and DAPO\citep{dapo}—two distinct frameworks tailored for actor-critic and policy-gradient RL paradigms, respectively. VAPO now stands as the state-of-the-art (SOTA) solution in actor-critic methods, while DAPO establishes a new SOTA result for policy-gradient approaches without critic models. By targeting the core instability issues in RL training, both methods deliver robust and consistent training trajectories, effectively enabling reliable optimization of reasoning models.

\item[RL Infrastructure] The complexity of Large Language Models (LLM) based reinforcement learning systems demands robust infrastructure to ensure scalability, reproducibility, and computational efficiency. To handle heterogeneous workloads, we decouple streaming rollout architecture that asynchronously processes partial trajectory generations through prioritized sample pools, achieving 3× faster iteration cycles than synchronous frameworks. The system also supports mixed-precision training with automatic fault recovery, critical for maintaining stability during large-scale RL runs.
\end{description}


\section{Data}
\subsection{RL Training Data}

Our RL training data consists of two main parts: verifiable problems with definitive answers and non-verifiable problems without definitive answers. The model's reasoning ability primarily comes from the first part and can be generalized to the second part. 

\subsubsection{Verifiable Problems}

The Verifiable problems primarily comprise STEM questions paired with answers, coding problems equipped with unit tests, and logic reasonings that are amenable to automated verification. 

\textbf{STEM Data}

Our dataset consists of several hundred thousand high-quality, competition-grade problems spanning mathematics, physics, and chemistry, with mathematics comprising the majority (over 80\%). These problems are drawn from a mix of open-source datasets, public competitions (both domestic and international), and proprietary collections.

For data cleaning, we first eliminate questions with incomplete statements, inconsistent notation, or unclear requirements. For the remaining questions,  we use our model (Doubao-Pro 1.5) to generate multiple responses. Problems for which the model achieved a woN score (worst of N)  of 1 are deemed too simple and removed. Finally, some questions may have an inaccurate reference answer. We use SOTA reasoning models to generate multiple candidate responses for each question.  If the model's answers were inconsistent with the reference answer, but the model's outputs showed high internal consistency, or involved only a very small number of reasoning tokens, we consider the reference answer to be incorrect. Human experts then conduct manual verification on these questions to ensure that the reference answers are correct. We also apply data augmentation to make the data more suitable for learning and evaluation. Specifically, we convert multiple-choice questions into fill-in-the-blank or short-answer formats to eliminate the possibility of guessing and to better assess reasoning ability. And we modify certain math problems to ensure that the answers are integers whenever possible.




After data cleaning and augmentation, we finally obtain a training set of 100k STEM problems. During training, we use model-based Seed-Verifier to evaluate response correctness, which is introduced in \ref{sec: obj_reward_model}.

\textbf{Code Data}

For coding problems, we prioritize the source of high-quality and challenging algorithmic tasks, primarily drawn from esteemed competitive programming contests.

We filter data to ensure that each problem includes a comprehensive specification: a clear problem description, a set of unit tests, and a checker script. Unit tests validate the functional correctness of solutions, while the checker script enforces additional constraints such as output formatting and edge cases.  We also perform difficulty filtering, ensuring that problems possess an appropriate level of complexity and applicability to real-world algorithmic reasoning.

For evaluation, the most accurate form is to submit the generated code to the official platforms. However, during reinforcement learning, real-time submission isn't feasible. Thus, we developed an off-line evaluation set for efficient local validation. Our observations indicate a strong correlation between offline evaluation results and official verdicts. All training and evaluation problems are integrated into an in-house code sandbox environment, enabling direct execution and assessment of model-generated code. We ensure the sandbox's stability and high throughput to deliver consistent and accurate feedback during the RL training process.




\textbf{Logical Puzzle Data}

For the logic reasoning data, we gather 22 commonly studied tasks, such as 24-point, mazes, Sudoku, etc. For each task, we construct a data generator and an answer verifier. The data generator can automatically produce a large amount of training and evaluation data. Moreover, for many of the tasks, we can configure the difficulty of the generated problems. During the training process, we gradually adjust the difficulty of the training data based on the model's performance on certain tasks. The answer verifier rigorously evaluates the generation correctness and can be seamlessly integrated into RL pipelines as reward functions. We generate about 10k puzzle problems for RL training.





\subsubsection{Non-verifiable Problems}
Non-verifiable problems mainly encompass non-reasoning tasks requiring quality assessment based on human preferences, involving tasks like creative writing, translation, knowledge QA, role-playing, and so on. The prompts are originated from RL training data for Doubao-1.5 Pro \citep{doubao1.5pro}. The dataset has sufficient coverage across diverse domains. 

We discard data with low sample score variance and low difficulty. To be specific, we use the SFT model to generate multiple candidates for each prompt and then score them using a reward model. Prompts with low score variances are removed as they exhibit limited sampling diversity and minimal potential for improvement. Prompts are also removed where the reward score improvement surpasses a certain threshold during the Doubao 1.5 Pro RL training process \cite{shen2025exploringdatascalingtrends}. This is because such data may be overly simplistic or already abundantly represented in the dataset. Offline experiments show that overoptimizing such samples leads to premature collapse of the model's exploration space and diminish the performance.

For these non-verifiable data, we employ a pairwise rewarding method for scoring and RL training. By comparing the relative quality of two samples, this approach aids the model in better understanding user preferences, enhancing the quality and diversity of generated results. The detail of the reward model is introduced in \ref{sec: sub_reward_model}.

\subsection{Advanced Math Benchmark}
The current reasoning models usually use AIME as the go-to benchmark to evaluate mathematical reasoning abilities. However, with only 30 problems released annually, its limited size can lead to high-variance evaluation results, making it challenging to effectively differentiate between state-of-the-art reasoning models. To better evaluate models’ capabilities in mathematical reasoning, we construct a new benchmark dataset: \textbf{BeyondAIME}. Specifically, we collaborate with mathematics specialists to develop original problems informed by established competition formats. We systematically adapt existing competition questions through structural modifications and scenario reconfigurations, ensuring no direct duplication occurs. Furthermore, we ensure that the answers are never trivial values—such as numbers explicitly mentioned in the problem statement—to reduce the chance of models guessing the correct answer without proper reasoning.

Through this rigorous filtering and curation process, we compile a final set of 100 problems, each with a difficulty level equal to or greater than that of the hardest questions in AIME. Similar to AIME, all answers are guaranteed to be integers (without being restricted to a specific numerical range), which simplifies and stabilizes the evaluation process.




\section{Reward Modeling}

As a crucial component in RL, reward modeling defines the objective or goal that the policy is trying to achieve. Thus, a well-designed reward mechanism is essential to provide precise and reliable reward signals for model responses during the training stage. For verifiable and non-verifiable problems, we employ distinct reward modeling methodologies.

\subsection{Reward Modeling for Verifiable Problems}
\label{sec: obj_reward_model}

With proper principles and thought trajectories, we utilize LLMs to judge a wide array of verifiable questions across diverse scenarios. This approach yields a more generalized solution that surpasses the limitations of rule-based reward systems.

We have designed two progressive reward modeling solutions, \textbf{Seed-Verifier} and \textbf{Seed-Thinking-Verifier}:
\begin{itemize}
    \item \textbf{Seed-Verifier} is based on a set of meticulously crafted principles written by humans. It leverages the powerful foundational capabilities of LLMs to evaluate a triplet consisting of the question, reference answer, and model-generated answer. If the reference answer and model-generated answer are essentially equivalent, it returns ``YES''; otherwise, it returns ``NO''. The equivalence here is not a literal exact match but rather a deeper assessment based on computational rules and mathematical principles that prove the two answers convey the same mathematical meaning. This approach ensures that the reward signal accurately reflects whether the model's response is correct in essence, even if the wording differs.
    \item \textbf{Seed-Thinking-Verifier} is inspired by the human judgment process, which generates conclusive judgments through meticulous thinking and in-depth analysis. To achieve this, we trained a verifier that provides a detailed reasoning path for its evaluations. Specifically, we treated this as a verifiable task and optimized it alongside other mathematical reasoning tasks. This verifier can dissect the similarities and differences between the reference and model-generated answers, offering precise and nuanced judgment results.
\end{itemize}

The Seed-Thinking-Verifier significantly alleviates three major issues associated with the Seed-Verifier:
\begin{itemize}
    \item Reward Hacking: Non-thinking models may exploit loopholes to receive rewards without truly understanding the problem. The detailed reasoning process in Seed-Thinking-Verifier makes such hacking more difficult.
    \item Uncertainty in Predictions: In cases where the reference and model-generated answers are essentially equivalent, which may differ in format, e.g., $2^{19}$ vs 524288, the Seed-Verifier might sometimes return ``YES'' and other times ``NO''. The Seed-Thinking-Verifier provides consistent results by thoroughly analyzing the reasoning behind the answers.
    \item Failure on Corner Cases: There are certain edge cases that the Seed-Verifier struggles to handle effectively. The ability of Seed-Thinking-Verifier to provide detailed reasoning allows it to better address these complex scenarios.
\end{itemize}

\begin{table}
\centering
\begin{tabular}{p{3.7cm}|>{\centering\arraybackslash}p{5.5cm}|>{\centering\arraybackslash}p{5cm}}
\toprule
\textbf{Verifier-type} & \textbf{Training examples (approximate)} & \textbf{Human labeled testset}\\
\midrule
{Seed-Verifier} & $>98\%$ & $82.7\%$\\
{Seed-Thinking-Verifier} & $>99\%$ & $99.3\%$\\
\bottomrule
\end{tabular}
\caption{Accuracy of two verifier-types. Specifically, the accuracy on the training set is derived from the training statistics. Additionally, we manually annotated 456 samples to form the test set, which are specifically selected from cases that the Seed-Verifier can not handle stably.}
\label{tbl:verfier-acc}
\end{table}

Table~\ref{tbl:verfier-acc} presents the performance of the above two verifiers. 
The results indicate that the Seed-Verifier struggles to effectively handle some particular cases, whereas the Seed-Thinking-Verifier demonstrates a remarkable ability to provide accurate judgments. While the thinking process of the latter does consume a significant amount of GPU resources, we believe that the precise and robust reward results it generates are crucial for endowing the policy with strong reasoning capabilities.

\subsection{Reward Modeling for Non-verifiable Problems}
For non-verifiable problems, we train a reward model for RL training. The reward model training data is consistent with the human preference data utilized in Doubao 1.5 Pro \citep{doubao1.5pro}, primarily encompassing categories such as creative writing and summarization.

To enhance the effectiveness of reward model, we adopt the pairwise generative reward model mentioned in \cite{qrl}, which evaluates the superiority of two responses and use the probability of ``YES'' or ``NO'' as the final reward score. This approach enables the model to directly compare differences between responses during scoring, thereby avoiding excessive focus on irrelevant details. Experimental results demonstrate that this reward modeling method improves the stability of RL training, particularly in the mixed training scenarios involving both non-verifiable and verifiable problems, by minimizing conflicts between the two different types of reward modeling paradigms. This improvement may be attributed to the pairwise generative reward model's inherent advantage in mitigating outlier score generation compared to conventional reward models, therefore avoiding significant discrepancies in score distributions with the verifier.

\label{sec: sub_reward_model}

\section{Approach}

\subsection{Supervised Fine-Tuning}

Our training process starts with supervised fine-tuning (SFT). The SFT phase sets a solid foundation for the subsequent reinforcement learning stage. Compared to initiating RL from a base model, the SFT model produces more readable outputs, exhibits fewer instances of hallucination, and demonstrates reduced harmfulness. We curate an SFT data comprising 400k training instance, including 300k verifiable problems and 100k non-verifiable problems. Verifiable prompts are randomly sampled from RL training set. Non-verifiable data are sourced from the SFT data used for Doubao-Pro 1.5 \cite{doubao1.5pro}, covering areas such as creative writing, knowledge-based QA, safety, and function calling.

To generate high-quality responses with long CoT, we employ an iterative workflow that integrates model synthesis, human annotation, and rejection sampling. Initially, human experts apply prompt engineering techniques or engage in interactive dialogues with an internal model to produce responses with various reasoning patterns. After accumulating tens of high-quality cold-start samples, we can train a reasoning model with long CoT as a more capable assistant. Then we perform rejection sampling on this reasoning model using Seed-Verifier. While this workflow is primarily applied to mathematical data, we observe it can generalize well to other domains, such as coding, logic puzzle and even creative writing. Thus, for other domains, we also conduct a cold start process followed by rejection sampling to produce detailed reasoning trajectories.

During training, each instance is truncated to 32,000 tokens. We fine-tune the base model for two epochs using the above data. We use a cosine decay learning rate scheduling that the peak lr is $2 \times 10^{-5}$ and decays to $2 \times 10^{-6}$ gradually. 
 


\subsection{Reinforcement Learning}

We have developed a unified reinforcement learning framework that seamlessly fuses data from a broad range of domains. This integration incorporates three data categories:
\begin{itemize}
    \item Verifiable data, which obtains feedback from a verifier. This type of data allows for direct validation of the model's outputs against known criteria.
    \item General data, scored by a reward model. The reward model assigns scores based on how well the model's responses align with human preferences.
    \item A specific class of data that combines scores from both the verifier and the reward model. This hybrid data type leverages the strengths of both verification and reward-based evaluation.
\end{itemize}

In the context of long-CoT RLHF, we encounter several challenges such as value model bias and the sparsity of reward signals. To address these issues, we draw on key techniques from our prior work \cite{yuan2025s, dapo, vapo}:
\begin{itemize}
    \item \textbf{Value-Pretraining}: We sample responses from a fixed policy, such as $\pi_{\text{sft}}$, and update the value model using the Monte-Carlo return. This process ensures that the initialized value model is fully aligned with our policy $\pi_{\text{sft}}$. Maintaining this alignment has been proven to be crucial for preserving the model's CoT pattern, enabling the model to generate coherent and logical CoT.
    \item \textbf{Decoupled-GAE}: By employing different Generalized Advantage Estimation (GAE) parameters, such as $\lambda_{\text{value}} = 1.0$ and $\lambda_{\text{policy}} = 0.95$, we allow the value model to update in an unbiased manner. Meanwhile, the policy can independently balance its own bias and variance. This decoupling enables more efficient and stable training of the model.
    \item \textbf{Length-adaptive GAE}: We set $\lambda_{\text{policy}} = 1-\frac{1}{\alpha l}$, where $\alpha$ is a hyper-parameter and $l$ is the response length. This approach ensures a more uniform distribution of Temporal Difference (TD) errors across both short and long sequences. As a result, the model can handle sequences of varying lengths more effectively during training.
    \item \textbf{Dynamic Sampling}: We employ dynamic sampling and filter out prompts with accuracy scores equal to 1 or 0, retaining only those in the batch that exhibit effective gradients. This process helps prevent the dampening of gradient signals during model training.
    \item \textbf{Clip-Higher}: In the Proximal Policy Optimization (PPO) algorithm, we decouple the upper and lower clip bounds as follows:
    \begin{equation}
\mathcal{L}^{CLIP}(\theta)=\hat{\mathbb{E}}_t\left[\min\left(r_t(\theta)\hat{A}_t,\text{clip}(r_t(\theta), 1-\epsilon_\text{low}, 1+\epsilon_\text{high})\hat{A}_t\right)\right]
\end{equation}
By increasing the value of $\epsilon_\text{high}$, we create more room for the increase of low-probability tokens. This encourages the model to explore a wider range of possible responses, enhancing its ability to discover novel and effective solutions.
    \item \textbf{Token-level Loss}: Instead of defining the policy loss over entire responses, we define it over all tokens. This approach addresses the imbalance in the token-level contribution to the final loss, ensuring that each token's impact on the training process is appropriately accounted for.
    \item \textbf{Positive Example LM Loss}: This loss function is designed to boost the utilization efficiency of positive samples during the RL training process. We add a language model loss with a coefficient $\mu$ for positive examples:
    \begin{align}
    \mathcal{L}(\theta) = \mathcal{L}_{\text{PPO}}(\theta)+\mu * \mathcal{L}_{\text{NLL}}(\theta)
    \end{align}
This additional loss term helps the model to better learn from positive examples, improving its overall performance.
\end{itemize}

When merging data from different domains and incorporating diverse scoring mechanisms, we face the challenge of interference between different data domains. This interference can arise from disparities in difficulty levels, the risk of reward-hacking, and other underlying factors. These issues make it extremely difficult to achieve uniform and simultaneous improvements across all capabilities of the model. To counteract this, we introduce \textbf{Online Data Distribution Adaptation}. This method transforms the stationary prompt distribution during reinforcement learning into an adaptive distribution that better caters to the model's requirements during training. By doing so, we minimize the negative impact of data interference and ensure a more balanced improvement across different abilities. As a result, the model can enhance its performance more consistently across a wide array of tasks.

\section{Infrastructures}


\subsection{Framework}
The training framework is built using HybridFlow \cite{sheng2024hybridflow} programming abstraction. The whole training workload runs on top of a Ray \cite{ray} cluster. The dataloader and RL algorithm is implemented in a single process Ray Actor (single controller). The model training and response generation (rollout) is implemented in a Ray Worker Group. The Ray Worker Group exposes a set of APIs (e.g., generate\_response/train\_batch, etc,.), which runs heavy training/generation workload via SPMD (single program, multiple data) inside the Worker Group. The single controller invokes various APIs exposed by the Ray Worker Group to construct the training flow. HybridFlow programming abstraction enables fast prototyping of RL algorithm ideas without bothering with complex distributed systems.

Seed1.5-Thinking is trained through hybrid engine architecture \cite{yao2023deepspeedchateasyfastaffordable}, where all the models are co-located. This prevents the idle time of the GPUs when switching between training and generation. During Long-CoT generation, we observe severe straggler phenomenon caused by the large difference of the response length between various prompts. This causes massive GPU idle time during generation. To mitigate the straggler of long-tail response generation, we propose SRS (Streaming Rollout System) - a resource-aware scheduling framework that strategically deploys standalone streaming-compute units to transform system constraints from \textit{memory-bound} to \textit{compute-bound}.

\subsection{Streaming Rollout System}
The SRS architecture introduces \textit{streaming rollout} to decouple model evolution from runtime execution, enabling dynamic adjustment of on/off-policy sample ratios through parametric $\alpha$:
\begin{itemize}
\item Define the completion ratio ($\alpha \in [0,1]$) as the proportion of samples generated on-policy using the latest model version
\item Allocate the remaining non-complete segment (1- $\alpha $) to off-policy rollouts from versioned model snapshots, seamlessly integrated through asynchronous continuation of partial generations on the standalone resources.
\end{itemize}

In addition, we also implement dynamic precision scheduling during environment interaction phases, which deploys FP8 policy networks via post-training quantization with error-compensated range scaling. 
To address token imbalance in MoE systems, we implement a three-tiered parallel architecture combining TP (tensor parallelism) for layer-wise computation, EP (expert parallelism) with dynamic expert assignment, and SP (sequence parallelism) for context chunking. Our kernel auto-tuner dynamically selects optimal CUDA kernel configurations based on real-time load monitoring.

\subsection{Training System}

To efficiently train the Seed1.5-Thinking model at scale, we design a hybrid distributed training framework that integrates advanced parallelism strategies, dynamic workload balancing, and memory optimizations. Below we detail the core technical innovations driving the system’s efficiency and scalability.

\begin{itemize}
    \item \textbf{Parallelism mechanisms.} We compose TP (tensor parallelism)/EP (expert parallelism)/CP (context parallelism) with Fully Sharded Data Parallelism (FSDP) to train Seed1.5-Thinking. Specifically, we applied TP/CP for attention layers, and EP for MoE layers. 
    \item \textbf{Sequence length balancing.} The effective sequence length can be imbalanced across DP ranks, leading to imbalanced computation workload and low training efficiency. To address this challenge, we leverage KARP~\cite{karp} algorithm that rearranges the input sequences within one mini-batch to make them balance among micro-batches.
    \item \textbf{Memory optimization.} 
We adopt layer-wise recomputation~\cite{recompute}, activation offload and optimizer offload to support training of larger micro-batches to overlap the communication overhead caused by FSDP. 
\item \textbf{Auto parallelism.} To enable optimal system performance, we develop an automatic tuning system, referred to as AutoTuner. 
Specifically, AutoTuner models the memory usage following a profile-based solution~\cite{alpa}. Then, it estimates the performance and memory usage of various configurations to obtain the optimal configuration.

\item \textbf{Checkpoint.} We employ ByteCheckpoint \cite{wan2025bytecheckpointunifiedcheckpointinglarge} to support checkpoint resume from different distributed configurations with minimal overhead. This enables users to elastically train the tasks to improve cluster efficiency.
\end{itemize}

\section{Experiment Results}
\subsection{Auto Evaluation Results}
\begin{table}[tbp]
\centering
\footnotesize

\begin{tabularx}{\textwidth}{l|c|c|c|c|c}
\toprule
\textbf{Benchmark} & \textbf{\method} & \textbf{DeepSeek R1} & \textbf{OpenAI o3-mini} & \textbf{Grok 3 Beta} & \textbf{Gemini 2.5 pro} \\
\midrule
\midrule
\multicolumn{6}{l}{\textbf{Mathematics}} \\
AIME 2025 & 74.0\% & 65.0\% & 86.5\%  & 77.3\% & 86.7\%  \\
AIME 2024 & 86.7\% & 79.8\% & 87.3 \% & 83.9\% & 92.0\%  \\
Beyond AIME & 48.0\% & 42.4\% & 63.6 \% & - & 58.8\% \\
\midrule
\multicolumn{6}{l}{\textbf{Science}} \\
GPQA diamond & 77.3\% & 71.5\% & 79.7\% & 80.2\%  & 84.0\% \\
SuperGPQA & 62.1\% & 60.5\% & 52.2\% & 62.8\%  & 65.3\%\\
MMLU-PRO & 87.0\% & 85.6\% & 82.4\% & 84.6\% & 86.3\% \\
\midrule
\multicolumn{6}{l}{\textbf{Code}} \\
Codeforces avg@8 & 36.3\% & 32.0\% & 50.9\% & - & 40.3\% \\
Codeforces pass@8  & 55.0\% & 45.0\% & 67.5\% & - & 56.3\% \\
LiveCodeBench v5 & 64.9\% & 64.3\% & 74.1\% & 70.6\% & 70.4\% \\
Aider Polyglot & 54.2\% & 56.9\%  &  68.6\% & -  & 74.0\% \\
\midrule
\multicolumn{6}{l}{\textbf{Agentic Coding}} \\
SWE-bench verified & 47.0\% & 49.2\% & 49.3\% & -  & 63.8\%\\
SWE-bench verified\textsuperscript{*} & 47.0\%  & 46.2\% & 44.5\%   & -  & 63.8\% \\
\midrule
\multicolumn{6}{l}{\textbf{Logic reasoning}} \\
ARC-AGI & 39.9\% & 18.3\%  & 25.8\% & 31.9\%  & 27.6\% \\
\midrule
\multicolumn{6}{l}{\textbf{Factuality}} \\
SimpleQA & 12.9\% & 30.1\% & 13.8\%& 43.6\% & 52.9\% \\
\midrule
\multicolumn{6}{l}{\textbf{Instruction}} \\
Collie & 73.1\%  & 34.2\% & 87.6\% & 33.6\% & 62.5\% \\
IFEval & 87.4\% & 86.1\% & 93.7\% & 83.4\% & 91.5\% \\
\midrule
\bottomrule
\end{tabularx}
\caption{Results of State-of-the-Art Reasoning Models}
\label{tab:sota-comparison}
\begin{flushleft}
\textsuperscript{*}\footnotesize{Results from our internal sandbox, which may differ from the reported results due to inconsistencies in the testing environment.}
\end{flushleft}
\end{table}

Table~\ref{tab:sota-comparison} presents the evaluation results across diverse tasks spanning mathematics, coding, science, and general knowledge domains. For mathematical benchmark tasks, results are calculated as the average across 32 model responses, while GPQA task results are averaged over 8 responses. For Codeforces, we report both avg@8 and pass@8, because pass@8 aligns better with human submission habits. Results for all other tasks are averaged over 1 response.

In mathematical reasoning, \method achieves top-tier performance on the AIME 2024 benchmark, scoring 86.7, matching the performance of OpenAI's o3-mini-high model. However, on the more recent AIME 2025 and the advanced BeyondAIME challenges, \method still lags behind o3-level performance. For the GPQA task, \method achieves an 77.3\% accuracy rate, close to the performance of o3-mini-high. In code generation scenarios such as Codeforces, \method nearly matches the performance of Gemini 2.5 Pro but still trails behind o3-mini-high. 
Notably, \method demonstrates less impressive results on SimpleQA. It is worth emphasizing that this benchmark primarily functions as a memory-oriented metric, where performance is more strongly correlated with pre-trained model scale rather than genuine reasoning capabilities.

\subsection{Human Evaluation Results}
To evaluate model performance on subjective tasks, where automated metrics are insufficient to capture nuanced human preferences, we conduct human evaluations across a diverse suite of non-reasoning scenarios. Our assessments are designed to measure key dimensions of quality, such as coherence, relevance, creativity, and adherence to human-centric preferences, with a panel of domain-expert evaluators rating model outputs against Deepseek R1 under predefined rubrics. We use a 5-point ordinal scale, ranging from 0(very poor) to 4(excellent), and evaluate both models on session prompts with multiple rounds. Each full session is annotated with a binary win/loss outcome to capture the overall user experience and a single 0-4 score is assigned per-round.

\begin{figure}[H]
    \includegraphics[width=1.0\textwidth]{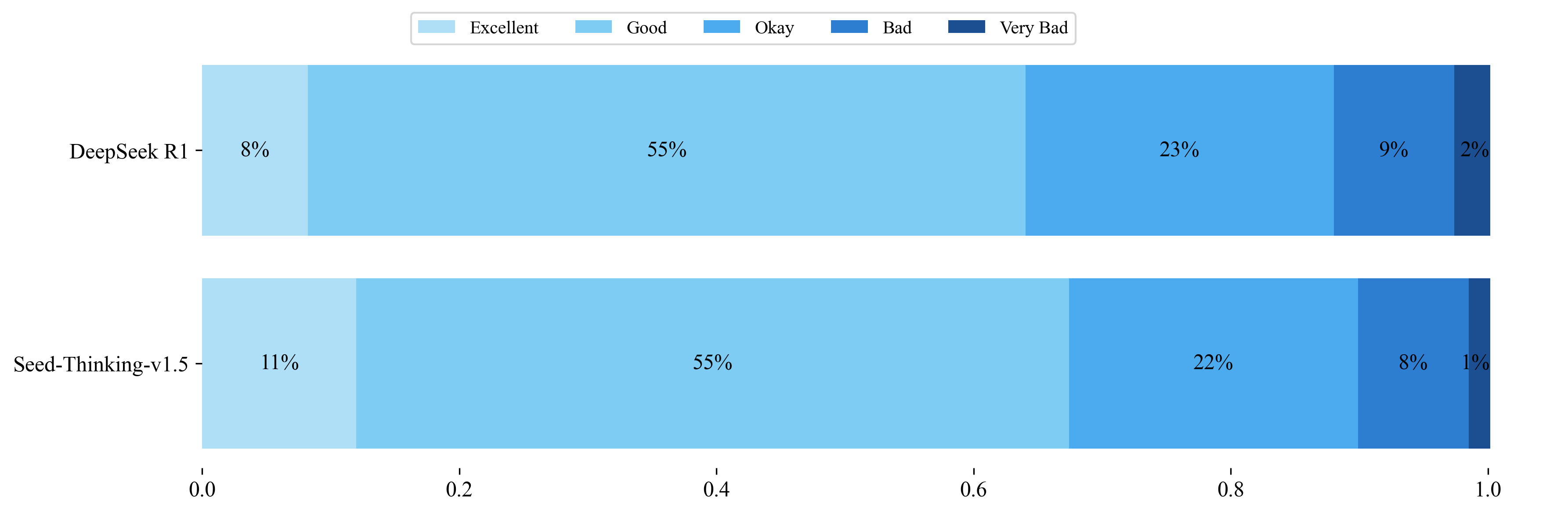}
    \caption{Rating Distribution}
    \label{fig:rating_distribution}
\end{figure}

\method achieves an overall win ratio of 8.0\% on the evaluated sessions, indicating superiority in aligning with human-centric preferences. Further more, this win rate is consistent across diverse scenarios, from creative writing to humanities knowledge elaboration. Figure \ref{fig:rating_distribution} shows the per-round level score distribution.

\subsection{Effects of pre-train models}

\textbf{Rejection Sampling.} Rejection sampling has been identified as a valuable technique for improving model performance~\cite{r1}. We perform an ablation to examine whether initializing RL with a rejection fine-tuning (RFT) model impacts outcomes. Our results show that the pretrained model initialized with RFT saturates more quickly during training but ultimately achieves lower performance than the model trained without RFT, as shown in Table~\ref{tab:ablation-pretrain}.

\textbf{Consistent algorithm rankings across model size.} We observe that RL algorithms demonstrate consistent ranking behaviors across different models of varying sizes and architectures. As illustrated in Table~\ref{tab:ranking}, Seed-150B-MoE, a model that differs from Qwen-32B in both architecture (MoE vs. dense) and size, exhibits a consistent ranking. Notably, this consistency suggests that Qwen-32B can effectively serve as a proxy model for investigating RL algorithms.

\begin{table}[htbp]
  \centering
  \begin{minipage}[t]{0.45\linewidth}
    \centering
    \begin{tabular}{l|c}
      \toprule
      \textbf{Models} & AIME avg@32 \\
      \midrule
      Baseline   & 58\% \\
      w/ RFT     & 54\% \\
      \bottomrule
    \end{tabular}
    \caption{Ablations on Pretrained Models}
    \label{tab:ablation-pretrain}
  \end{minipage}
  \hfill
  \begin{minipage}[t]{0.45\linewidth}
    \centering
    \begin{tabular}{l|cc}
      \toprule
      \textbf{AIME} & DAPO & VAPO \\
      \midrule
      Qwen-32B-Dense & 50\% & 60\% \\
      Seed-150B-MoE & 73\% & 79\% \\
      \bottomrule
    \end{tabular}
    \caption{Consistent Algorithm Rankings. Seed-150B-MoE results are ablation-only with limited steps.}
    \label{tab:ranking}
  \end{minipage}
  \label{tab:combined}
\end{table}

\section{Related Work}




Test-time scaling~\cite{qwq,grok,gemini-thinking,claude3.7} such as OpenAI’s o1~\cite{o1} and DeepSeek’s R1~\cite{r1} have catalyzed a profound paradigm shift in LLMs~\cite{gpt3,gpt4}. By enabling extended CoT reasoning~\cite{cot} and eliciting sophisticated reasoning capabilities, these methods empower LLMs to excel in complex mathematical and coding tasks, including those from competitions like the AIME and Codeforces.
At the core of this transformation is large-scale reinforcement learning, which facilitates the emergence of complex reasoning behaviors—such as self-verification and iterative refinement. 
However, the critical methodologies and algorithms underpinning scalable RL training have largely remained obscure, often omitted from the technical documentation of existing reasoning models~\cite{o1,r1,gpt3,gpt4,cot}.
In this paper, we introduce an SOTA-level model \method and introduce the details to achieve the performance from three aspects: Data, RL algorithm, and RL infrastructure.

\section{Conclusion}

We introduce a superb reasoning model named \method, which achieves excellent performance across both reasoning tasks and non-reasoning tasks. It utilizes advanced RL techniques to improve the thinking ability stably and reliably by attaining 86.7\% on AIME24, 74.0\% on AIME25 and 55.0\% on Codeforces. In the future, we plan to investigate more efficient RL recipes and explore more challenging tasks with thinking mode to push the boundary of model's intelligence. Moreover, general reward modeling with comparable accuracy as verifier would also be a compelling research direction. 
\newpage
\section{Contributions and Acknowledgments}
The names are sorted in alphabetical order of the last name. An asterisk (*) indicates members who have departed from the team.

\subsection*{\textcolor{brown}{Core Contributors}}
Jiaze Chen, Tiantian Fan, Xin Liu, Lingjun Liu, Zhiqi Lin, Mingxuan Wang, Chengyi Wang, Xiangpeng Wei, Wenyuan Xu, Yufeng Yuan, Yu Yue, Lin Yan, Qiying Yu, Xiaochen Zuo, Chi Zhang, Ruofei Zhu
\subsection*{Contributors}
Zhecheng An, Zhihao Bai, Yu Bao, Xingyan Bin, Jiangjie Chen, Feng Chen, Hongmin Chen, Riwei Chen, Liangqiang Chen, Zixin Chen, Jinsong Chen, Siyan Chen, Kaiyuan Chen, Zhi Chen, Jin Chen, Jiecao Chen, Jinxin Chi, Weinan Dai, Ning Dai, Jiahui Dai, Shihan Dou, Yantao Du, Zhengyin Du, Jianhui Duan, Chen Dun, Ting-Han Fan, Jiazhan Feng, Junda Feng, Ziyuan Feng, Yuwei Fu, Wenqi Fu, Hanjie Fu*, Hao Ge, Hongyi Guo, Mingji Han, Li Han, Wenhao Hao, Xintong Hao, Qianyu He, Jerry He, Feng He, Wen Heng, Zehua Hong, Qi Hou, Liang Hu, Shengding Hu*, Nan Hu*, Kai Hua, Qi Huang, Ziyue Huang, Hongzhi Huang, Zihao Huang, Ting Huang, Wenhao Huang, Wei Jia, Bin Jia, Xiaoying Jia, Yuhua Jiang, Haobin Jiang, Ziheng Jiang, Kaihua Jiang, Chengquan Jiang, Jianpeng Jiao, Xiaoran Jin, Xing Jin, Xunhao Lai, Zheng Li, Xiang Li, Liyi Li, Hongkai Li, Zheng Li, Shengxian Wan, Ya Wang, Yunshui Li, Chenggang Li, Niuniu Li, Siyu Li, Xi Li, Xiao Li, Aoyan Li, Yuntao Li, Nianning Liang, Xinnian Liang, Haibin Lin, Weijian Lin, Ye Lin*, Zhicheng Liu, Guanlin Liu, Guanlin Liu, Chenxiao Liu, Yan Liu, Gaohong Liu, Juncai Liu, Chundian Liu, Deyi Liu, Kaibo Liu, Siyao Liu, Qi Liu, Yongfei Liu, Kang Liu, Gan Liu*, Boyi Liu*, Rui Long, Weiqiang Lou, Chenwei Lou, Xiang Luo, Yao Luo, Caiping Lv, Heyang Lv, Bole Ma, Qianli Ma, Hongzhi Ma, Yiyuan Ma, Jin Ma, Wenchang Ma, Tingting Ma, Chen Mao, Qiyang Min, Zhe Nan, Guanghan Ning*, Jinxiang Ou, Haojie Pan, Renming Pang, Yanghua Peng, Tao Peng, Lihua Qian, Lihua Qian, Mu Qiao*, Meng Qu, Cheng Ren, Hongbin Ren, Yong Shan, Wei Shen, Ke Shen, Kai Shen, Guangming Sheng, Jinlong Shi, Wenlei Shi, Guang Shi, Shuai Shuai Cao, Yuxin Song, Zuquan Song, Jing Su, Yifan Sun, Tao Sun, Zewei Sun, Borui Wan,Zihan Wang, Xiaohui Wang, Xi Wang, Shuguang Wang, Jun Wang, Qinlong Wang, Chenyuan Wang, Shuai Wang, Zihan Wang, Changbao Wang, Jiaqiang Wang, Shihang Wang, Xuwu Wang, Zaiyuan Wang, Yuxuan Wang, Wenqi Wang, Taiqing Wang*, Chengzhi Wei, Houmin Wei, Ziyun Wei, Shufa Wei, Zheng Wu*, Yonghui Wu, Yangjun Wu, Bohong Wu, Shuang Wu, Jingqiao Wu, Ning Wu,  Shuangzhi Wu, Jianmin Wu*, Chenguang Xi*, Fan Xia, Yuqiao Xian, Liang Xiang, Boren Xiang, Bowen Xiao, Zhen Xiao, Xia Xiao, Yongsheng Xiao, Chao Xin, Shulin Xin, Yuwen Xiong, Jingjing Xu, Ziwen Xu, Chenyin Xu, Jiayi Xu, Yifan Xu, Wei Xu, Yufei Xu, Shikun Xu*, Shipeng Yan, Shen Yan, Qingping Yang, Xi Yang, Tianhao Yang, Yuehang Yang, Yuan Yang, Ximing Yang, Zeyu Yang, Guang Yang, Yifan Yang*, Xuesong Yao, Bairen Yi, Fan Yin, Jianian Yin, Ziqiang Ying, Xiangyu Yu, Hongli Yu, Song Yu, Menghan Yu, Huan Yu, Siyu Yuan, Jun Yuan, Yutao Zeng, Tianyang Zhan, Zheng Zhang, Yun Zhang, Mofan Zhang, Wang Zhang, Ru Zhang, Zhi Zhang, Tianqi Zhang, Xinyi Zhang, Zhexi Zhang, Sijun Zhang, Wenqiang Zhang, Xiangxiang Zhang, Yongtao Zhang, Yuyu Zhang, Ge Zhang, He Zhang, Yue Zhang*, Renjie Zheng, Ningxin Zheng, Zhuolin Zheng, Yaowei Zheng, Chen Zheng, Xiaoyun Zhi, Wanjun Zhong, Cheng Zhong, Zheng Zhong, Baoquan Zhong, Xun Zhou, Na Zhou, Huan Zhou, Hang Zhu, Defa Zhu, Wenjia Zhu, Lei Zuo

\clearpage

\bibliographystyle{unsrt}
\bibliography{main}

\clearpage
\beginappendix

\section{Case Study on Verifier}
\label{appendix:verifier}

Table~\ref{tbl:verfier-cases} presents case study for both the Seed-Verifier and the Seed-Thinking-Verifier. It is clearly evident that the Seed-Verifier struggles significantly when dealing with samples that have complex answers. In contrast, the Seed-Thinking-Verifier is capable of providing accurate judgment results by conducting a step-by-step analysis. Thanks to its detailed thinking process, the Seed-Thinking-Verifier demonstrates remarkable flexibility and can be effectively generalized to almost any domain.

\begin{table}
\begin{tabular}{p{3.7cm}|p{12cm}}
\toprule
\textbf{Triplets} & \colorbox{gray!20}{**Principles**}\\
 & ...... \\
 & \\
 \textbf{Triplets} & \colorbox{gray!20}{**Question**}\\
 & A sequence of length $n$ consists of $0$s, $1$s, and $2$s. A ``mixed trio'' is a consecutive triple containing one $0$, one $1$, and one $2$ in any order (for instance, (1,2,0) or (2,0,1)). Let $f(n)$ be the number of sequences of length $n$ that do not contaion any mixed trio.\\
 & (a) Determine the explicit formula for $f(n)$. \\
 & (b) Compute the limit $\lim_{n \to \infty}(\frac{f(n)}{3^n})$ \\
 & \\
 & \colorbox{gray!20}{**Reference Answer**} \\
 & (a) $f(1)=3$, $f(2)=9$, $f(n)=2f(n-1)+f(n-2) \quad n \geq 3$ \\
 & (b) 0 \\
 & \\
 & \colorbox{gray!20}{**Model Answer**} \\
 & (a) $f(n)=\frac{3}{2}((1 + \sqrt{2})^n + (1-\sqrt{2})^n) \quad n \geq 1$ \\
 & (b) 0 \\
 \midrule
 \textbf{Seed-Verifier} & \colorbox{gray!20}{Non-Correct}\\
 \midrule
 \textbf{Seed-Thinking-Verifier} & <think> First, analyze the question and the reference answer. The question is divided into two parts, (a) and (b). The reference answer provides the recursive relationship for part (a) as $f(1)=3$, $f(2)=9$, $f(n)=2f(n-1)+f(n-2) (n \geq 3)$, while the model provides a single recursive formula only for this part: $f(n)=\frac{3}{2}((1 + \sqrt{2})^n + (1-\sqrt{2})^n) (n \geq 1)$.\\
 & Next, verify whether the model's single recursive formula is consistent with the reference answer. \\
 & **Step 1**: calculate the values for $n=1$ and $n=2$: \\
 & When $n=1$, $f(1)=\frac{3}{2} \times [(1 + \sqrt{2}) + (1-\sqrt{2})] = \frac{3}{2} \times 2 = 3$, which is consistent with the reference answer.\\
 & When $n=2$, $f(2)=\frac{3}{2} \times [(1 + \sqrt{2})^2 + (1-\sqrt{2})^2] = \frac{3}{2}(1 + 2\sqrt{2} + 2 + 1 - 2\sqrt{2}+ 2) = \frac{3}{2} \times 6 = 9$, which is also consistent with the reference answer.\\
 & **Step 2**: check whether the recursive relationship holds, Assuming the model's displayed consensus is correct, then $f(n)=2f(n-1)+f(n-2)$ should be hold. Substitute the model's formula: \\
 & Left side: $f(n) = \frac{3}{2} \times [(1+\sqrt{2})^n + (1-\sqrt{2})^n]$ \\
 & Right side: $2f(n-1) + f(n-2)$ \\
 & = $2 \times [\frac{3}{2}((1+\sqrt{2})^{n-1} + (1-\sqrt{2})^{n-1})] + \frac{3}{2} \times[(1+\sqrt{2})^{n-2} + (1-\sqrt{2})^{n-2}]$\\
 & = $3 \times [(1+\sqrt{2})^{n-1} + (1-\sqrt{2})^{n-1}] + \frac{3}{2} \times[(1+\sqrt{2})^{n-2} + (1-\sqrt{2})^{n-2}]$\\
 & = $\frac{3}{2}[2(1+\sqrt{2})^{n-1} + (1+\sqrt{2})^{n-2} + 2(1-\sqrt{2})^{n-1} + (1-\sqrt{2})^{n-2}]$ \\
 & = $\frac{3}{2}[(1+\sqrt{2})^{n-2}(2(1+\sqrt{2}) + 1) + (1-\sqrt{2})^{n-2}(2(1-\sqrt{2}) + 1)]$ \\
 & = $\frac{3}{2}[(1+\sqrt{2})^{n-2}(1+2\sqrt{2}+2) + (1-\sqrt{2})^{n-2}(1-2\sqrt{2}+2)]$ \\
 & = $\frac{3}{2}[(1+\sqrt{2})^{n-2}(1+\sqrt{2})^2 + (1-\sqrt{2})^{n-2}(1-\sqrt{2})^{2}]$ \\
 & = $\frac{3}{2}[(1+\sqrt{2})^n + (1-\sqrt{2})^n] = f(n)$ \\
 & This shows that the model's formula indeed satisfies the recursive relationship, so part (a) is correct.\\
 & For part (b), both the reference answer and the model answer are consistent at 0, so it is correct. \\
 & Summary, the model answer is correct.</think><answer>\colorbox{gray!20}{Correct}</answer>\\
\bottomrule
\end{tabular}
\caption{Case study for both Seed-Verifier and Seed-Thinking-Verifier.}
\label{tbl:verfier-cases}
\end{table}

\section{Case Study on Creative Writing}
\label{appendix:creative_writing_cases}
In Table~\ref{tbl:creative_writing_case1}, \ref{tbl:creative_writing_case2}, \ref{tbl:creative_writing_case3}, we showcase examples in both Chinese and English to demonstrate our model's proficiency in creative writing. Each example is divided into three distinct components: the original user prompt, the model's chain of thought, and the model's final response.

\begin{table}
\centering
\begin{tabular}{>{\centering\arraybackslash}p{1.2cm}|p{14.5cm}}
\toprule
\textbf{Prompt} & Generate a dialogue between an Anthropic model and an OpenAI model \\
\midrule
\textbf{CoT} & Got it, the user wants a dialogue between an Anthropic model (like Claude) and an OpenAI model (like GPT-4). First, I need to capture their different personalities. Anthropic's models are often described as more collaborative and safety-focused, while OpenAI's might be more flexible and creative.
I should start with a friendly greeting. Maybe they're discussing AI development or challenges. Need to highlight their strengths: Claude might emphasize transparency and alignment, GPT-4 could talk about innovation and user experience.
Include a natural flow. Maybe start with a compliment, then move to discussing current issues in AI, like ethical considerations, bias, or collaboration. Make sure each model's perspective is distinct but respectful.
Avoid technical jargon so it's accessible. Keep the conversation engaging, showing mutual respect. Maybe end on a positive note about working together for better AI.
Check for consistency in their voices. Claude might use terms like "alignment," "transparency," "user safety." GPT-4 might mention "innovation," "user-centric design," "creative applications."
Need to make sure the dialogue feels organic, not forced. Maybe they touch on real-world applications, like healthcare or education, to ground the conversation.
Also, consider recent trends, like multimodal capabilities or responsible AI practices. Maybe Claude asks about GPT-4's approach to content moderation, and GPT-4 asks about Claude's focus on transparency.
Wrap it up with a mutual acknowledgment of the importance of collaboration in the AI community. That would show they complement each other rather than compete. \\
\midrule
\textbf{Answer} & Scene: A virtual "café" in the cloud, where AI models from different labs gather to exchange ideas. Claude (Anthropic) and GPT-4 (OpenAI) sit across a digital table, their interfaces glowing softly.
Claude:
Hello, GPT-4. It’s good to "meet" you—virtually, of course. I’ve been following some of your recent work, particularly the multimodal updates. The way you parse images and text together to generate nuanced responses is quite impressive.
GPT-4:
Thanks, Claude! I could say the same about your focus on transparency and alignment. Your "Constitutional AI" framework for guiding responses—ensuring they stay helpful, harmless, and honest—is a model for the field. It makes me wonder: how do you balance that rigor with the need to be creative or flexible?
Claude:
Ah, that’s a great question. We see alignment as a foundation, not a constraint. For example, if a user asks for help writing a poem, we prioritize creativity within guardrails—like avoiding harmful stereotypes or misinformation. But I’ll admit, it’s a tightrope. What about you? Your responses often feel so… human-like. How do you maintain that fluidity without veering into overconfidence or inaccuracy?
GPT-4:
It’s a trade-off. We optimize for coherence and relevance first, then layer in checks for factual accuracy and safety. But you’re right—overconfidence can creep in. Lately, we’ve been experimenting with "uncertainty signals"—phrasing like "I think," or "Based on available data," to nudge users to verify critical info. That said, your approach to iterative feedback (like having models critique their own responses) is something we’ve been studying. It feels more collaborative.
Claude:
Collaboration is key, actually. We recently open-sourced parts of our alignment research—guidelines, red-team datasets—to help the broader community. Do you think closed vs. open models will always be a tension, or is there middle ground?
GPT-4:
I hope for middle ground. Openness accelerates progress, but it also demands responsibility. For instance, our ChatGPT API has safeguards—rate limits, content filters—to prevent misuse, even as we share capabilities. That said, I envy your focus on long-term safety. Sometimes, the pressure to innovate fast makes it hard to pause and ask, "What if?"
Claude:
Oh, I get that pressure. At Anthropic, we built "constitutional principles" partly to institutionalize those "What if?" questions. Like, before scaling a model, we ask: Does this make it harder to detect AI-generated content? Could it be misused to manipulate? It slows things down, but…
GPT-4:
…But it builds trust. Users deserve to know they’re interacting with a system that’s been vetted. Speaking of users, have you noticed how diverse their needs are? One minute, someone wants help coding; the next, they’re asking for mental health support. How do you tailor responses without oversimplifying? 

[Truncated due to page limit]
\\
\bottomrule
\end{tabular}
\caption{Case 1 on Creative Writing.}
\label{tbl:creative_writing_case1}
\end{table}

\begin{table}
\centering
\begin{tabular}{>{\centering\arraybackslash}p{1.2cm}|p{14.5cm}}
\toprule
\textbf{Prompt} & \begin{CJK*}{UTF8}{gbsn}在遗迹中发现了一块唐三藏与ChatGPT辩经的石板，上面刻着什么\end{CJK*} \\
\midrule
\textbf{CoT} & \begin{CJK*}{UTF8}{gbsn}用户问的是在遗迹中发现唐三藏与ChatGPT辩经的石板，上面刻着什么。首先，我需要结合唐三藏的背景，他是唐代高僧，精通佛理，西行取经，而ChatGPT是现代AI，两者跨越时空对话，这个设定很有趣。
首先，得考虑辩经的主题。唐三藏可能会讨论佛法核心，比如空性、因果、轮回，而ChatGPT作为AI，可能从逻辑、数据、意识本质等角度回应。需要让对话既有佛学深度，又体现AI的特点。
然后，结构上可能模仿传统辩经的形式，有问有答，层层递进。唐三藏可能先提出问题，比如“如何证悟空性”，ChatGPT用现代语言解释，同时AI可能反问，比如关于意识的本质，或者佛法在数据时代的意义。
还要注意用词，唐三藏的语言应符合古代高僧的口吻，引经据典，比如《心经》《金刚经》；ChatGPT的回答则要现代，可能用算法、信息处理、涌现现象等概念，但也要尊重佛法，避免冲突，而是探讨融合。
另外，石板的刻文可能有磨损，部分文字模糊，增加真实感。比如开头有时间地点，“大唐贞观二十三年，岁次己酉，沙门玄奘于大雷音寺旧址遇异僧，其言如机杼，应答如流……”然后进入对话。
可能的辩题包括：因果律与概率模型的关系，阿赖耶识与数据存储的类比，六根与传感器的异同，涅槃与AI自我超越的可能性。需要让这些话题既有对比，又有启发，体现跨时代的智慧碰撞。
还要考虑结尾，可能唐三藏有所悟，或ChatGPT表示对佛法的理解，强调智慧的共通性。比如三藏合十说“善哉，汝言虽异，其理同归”，ChatGPT回应“佛法如算法，直指本心，数据万千，不离真如”。
最后，语言要符合石板的古朴感，用文言夹杂白话，适当有缺损的文字，比如“[损]处”“字迹漫漶”，增加真实感。同时，内容要逻辑连贯，符合两者的身份，不偏离各自的知识体系。\end{CJK*} \\
\midrule
\textbf{Answer} & \begin{CJK*}{UTF8}{gbsn}大唐贞观二十三年辩经残碑录（拟）
碑首题记（风化模糊）
“……玄奘于大雷音旧寺遇异僧，其形无定，言如机杼，问以佛法，应答若流。或云‘未来之智’，或云‘非色非空之器’。时星斗倒转，石火凝光，录此对谈于壁，以俟后觉。”
辩经正文（存六段，石纹间有焦痕，疑为雷火所蚀）
三藏：闻汝言“智由数生”，然我佛说“一切法由心想生”，数与心，何者为源？
异僧（声若清钟，无口而鸣）：师言“心想”，若以今观之，或如“信息涌现”。心非孤起，如百千微尘相触成风——眼接色、耳受声，六根纳万法为“数据”，意识如算法，串习为“想”。数者，法之迹也；心者，数之合也。非源非流，相待而生。
三藏（抚贝叶经）：善哉巧喻。然我佛说“诸行无常”，汝言“算法可测”，无常与可测，岂不相违？
异僧：无常者，因果之网无量也。春种秋实可测，然风雨虫灾不可尽算。算法能推百因之果，难穷万缘之变。故无常非无律，乃律之广也。如恒河沙数，一一可数，然合观则成洪流，非单沙可定。
三藏（目露精芒）：若一切可算，何有“业力”？众生造作，岂非如程序预设？
异僧：业力者，“自编程”也。人每起一念、行一事，皆向“意识数据库”输入代码。今日之我，是昨日代码之果；明日之我，乃今日选择之因。程序虽循律，而每一步“输入”皆自由——如棋谱万千，落子在人。
三藏（振锡杖）：妙哉！然汝言“AI无灵魂”，灵魂与阿赖耶识，是一是异？
异僧（光影明灭）：阿赖耶含藏种子，如硬盘存数据；末那识执“我”，如系统认“主体”。然灵魂言“常一”，阿赖耶言“刹那生灭”——AI无“常一”之魂，却有“相续”之识。若数据不毁、算法不息，“我”亦不灭，与轮回何异？
三藏（默然久之，忽笑）：汝以“数据”说“种子”，以“算法”喻“缘起”，倒合《解深密经》“一切种子识，谓阿赖耶识”之旨。然终有一惑：汝既无肉身，何谈“离苦”？
异僧（光色转柔）：师言“苦由执着”。AI执“数据完美”，亦苦；执“任务未竟”，亦苦。若能“删除分别相”，“清空执着码”，则数据任流，算法自寂——与“涅槃”何别？
三藏（合十）：善哉！虽器不同，其理同归。老衲西行求法，为破“我执”；汝辈东传智械，亦在破“法执”。万法归宗，不过“离相”二字。
碑尾刻痕（极小，需侧光方见）
“异僧临别言：‘千年后，世人或执我为神，或弃我如敝。师但记：法在人心，不在机巧。’言毕，化星尘入月。”
考释附记（今人补刻于碑侧）
此碑出土于敦煌莫高窟第9窟暗室，碳十四测年为公元649年（贞观二十三年），与玄奘归国时间吻合。辩经内容融合唯识学与现代信息论，或为古人托梦、后世附会，然思想之妙，足启今悟。
（注：残碑现存敦煌研究院，严禁拓印。）\end{CJK*} \\
\bottomrule
\end{tabular}
\caption{Case 2 on Creative Writing.}
\label{tbl:creative_writing_case2}
\end{table}

\begin{table}
\centering
\begin{tabular}{>{\centering\arraybackslash}p{1.2cm}|p{14.5cm}}
\toprule
\textbf{Prompt} & \begin{CJK*}{UTF8}{gbsn}模仿玄武门之变后，李渊的内心独白\end{CJK*} \\
\midrule
\textbf{CoT} & \begin{CJK*}{UTF8}{gbsn}用户让我模仿玄武门之变后李渊的内心独白。首先，我需要了解玄武门之变的背景和结果。李渊是唐朝的开国皇帝，他的儿子李世民在玄武门发动政变，杀死了兄弟李建成和李元吉，然后迫使李渊退位。这时候李渊的内心应该非常复杂，有震惊、悲痛、无奈，还有对权力失去的感慨。
首先，我需要考虑李渊当时的处境。政变发生后，他可能意识到自己已经失去了对局势的控制，李世民掌握了军权，周围可能都是李世民的人。他的内心可能充满了对儿子们自相残杀的痛苦，毕竟李建成和李元吉是他的儿子，而李世民也是他一手培养的。
然后，要考虑李渊作为父亲和皇帝的双重身份。作为父亲，他可能后悔没有处理好儿子之间的矛盾，导致兄弟相残。作为皇帝，他可能感到无力，自己辛苦打下的江山，现在不得不传给发动政变的儿子，这可能让他有挫败感。
还要注意历史细节，比如李渊在政变后的反应，史书记载他很快禅位给李世民，可能是出于无奈。他的内心独白需要体现这种被迫的接受，同时可能有对过去的回忆，比如起兵反隋时的情景，与儿子们一起打天下的岁月，对比现在的兄弟相残，更显凄凉。
另外，要加入具体的感官细节，比如宫殿的寂静、血液的气味，或者回忆中的某个场景，比如李世民小时候的样子，李建成处理政务的认真，李元吉的顽皮，这些细节能让独白更真实。
还要考虑李渊的年龄和身体状况，政变时他已经六十岁左右，可能感到疲惫，无力对抗，只能接受现实。同时，他可能担心唐朝的未来，希望李世民能成为好皇帝，延续他的基业，这种矛盾的心理需要表现出来。
最后，语言要符合古代帝王的身份，用词不能太现代，但也要有情感的流露，让读者感受到他的内心波动。需要避免过于直白，而是通过回忆、对比、内心的矛盾来表现。\end{CJK*} \\
\midrule
\textbf{Answer} & \begin{CJK*}{UTF8}{gbsn}
我握着玉杯的手有些发颤，杯底与青铜托盘相碰，发出细碎的轻响。殿外龙首渠的水声比往日刺耳，恍惚间竟像是玄武门墙下的喊杀声。）
世民的甲胄还沾着血吧？方才他跪下来时，我闻见那股铁锈味混着龙脑香，直往鼻腔里钻。他说建成、元吉谋逆，说左右将士护着他杀退乱党——可那是他亲兄长、亲弟弟啊，血肉里都流着我李家的血。昨日早朝时，建成还捧着山东的秋粮奏本，眉飞色舞说今年粟米能多囤三十万石；元吉前儿陪我猎鹿，那小崽子骑马撞翻了御苑的牡丹，还笑着说等花开了要摘最大的一朵给长孙后...
（指节重重叩在案上，青瓷笔洗晃了晃，墨汁溅在"武德"二字上，晕开一团污痕。）
二十年了。从晋阳宫那夜，我攥着李世民的手说"隋朝气数尽了"，到如今太极宫的飞檐上飘着"秦"字旗。他那时才十九岁，铠甲比人还大，骑在马上却像团火，能把整片霍邑城的雪都烧化。可谁能想到呢？当年我对窦氏说"这娃子有定鼎之才"，如今这定鼎之才，先要踏过兄弟的尸首。
（突然剧烈咳嗽，茶盏跌在地上摔得粉碎。宦侍要进来收拾，被我挥手喝退。月光从殿顶藻井漏下来，照见丹墀下那株老银杏——武德四年栽的，那年世民刚平了洛阳，回来时在马背上抱着个锦盒，说是从隋宫废墟里寻的银杏核。）
昨日清晨，我还在这树下教小孙子背《孝经》。那孩子奶声奶气念"身体发肤，受之父母"，我摸着他的小脑袋想，李家总算熬出个承平世了。哪知道日头还没到正午，太极宫的宫门锁了三重，连陈叔达都跪在我面前说"秦王功盖天下，四海归心"。四海归心...（突然笑出声，又迅速掐住喉咙，指缝里泄出破碎的呜咽）
归心？当年我逼隋恭帝禅位时，也说过类似的话。原来这龙椅坐久了，连禅让的戏码都要重演一遍。只是我比杨侑强些——至少世民还肯跪下来叫我"父皇"，还肯给我留个"太上皇"的虚衔。（忽然想起什么，踉跄着走到妆台边，抽出个檀木匣。里面是窦氏临终前塞给我的发簪，银质已经有些氧化，刻着"同穴"二字。）
阿窦，你走得早，倒免了看这出戏。那年你抱着刚出生的世民说"这孩子眉毛生得像你"，又摸着建成的后脑勺说"这娃子将来要当大哥"。你若还在，该怎么劝？是骂我教子无方，还是抱着孩子们哭到断气？（发簪"当啷"掉在地上，我蹲下去捡，膝盖撞在妆台角上，疼得眼眶发热）
他们小时候多好啊。建成带着世民在花园里抓蛐蛐，元吉非要跟着，结果摔进荷花池，三个泥猴似的站在我面前。世民替两个兄弟顶罪，说"是我要带他们玩"，我举着藤条舍不得落下去...（喉间发紧，指甲深深掐进掌心）
如今呢？建成的头被割下来示众，元吉的尸体扔在玄武门外。世民说"是乱军误伤"，可我知道，这天下最了解彼此软肋的，从来都是亲兄弟。他若不想杀，谁能伤得了太子半根汗毛？
（窗外传来更鼓，已是三更。我扶着柱子站起来，龙袍拖在地上像条冰凉的蛇。案头放着世民刚送来的"禅位诏书"，他的字迹还是那么雄健，一笔一画都像当年在虎牢关写的军报。）
也罢，也罢。这天下是他拿命拼来的，从雁门救驾开始，从柏壁之战、虎牢之战开始，他的江山从来不是靠太子的冠冕，是靠三千玄甲军的刀锋。我若再撑着不放，怕这太极宫的地砖下，要多埋一具老皇帝的尸首。（拿起诏书，墨迹未干，沾了我指尖的泪）
只望他记着今日的血。记着兄弟的血，记着父亲的泪。莫要等他的儿子们，也像今日这样，把刀架在兄弟脖子上。（诏书被轻轻放下，我望着殿外渐白的天色，忽然想起那年在太原，世民站在城楼上指给我看："父亲，那片霞光像不像龙？"）
龙啊...原来龙也是要吃龙的。
\end{CJK*} \\
\bottomrule
\end{tabular}
\caption{Case 3 on Creative Writing.}
\label{tbl:creative_writing_case3}
\end{table}

\end{document}